\documentclass[review]{fcs}

\usepackage{times}% : 这个包设置了全文的字体为Times Roman。

\usepackage{graphicx}% : 这个包提供了一组强大的命令用于处理图形文件。

\usepackage{amsmath}% : 这个包提供了多种数学符号和公式环境的增强。
\usepackage{amssymb}% : 这个包增加了多种数学符号的支持。

\usepackage{hyperref}% : 这个包用于处理文档中的超链接。

\usepackage{bbm}% : 这个包可能用于产生黑板粗体字符，例如表示实数集的\mathbb{R}。

\usepackage{algorithm2e}% : 这个包可能用于在文档中插入算法。

\usepackage{multirow}% : 这个包可能用于创建跨多行的表格单元格。

\usepackage{amsmath,amsfonts,bm}% : 这些包用于提供数学公式和符号的支持。

\usepackage{cleveref}% : 这个包用于更智能地引用其他部分（如图、表、公式等）。
\usepackage{booktabs}
\usepackage{makecell}
\usepackage{comment}
\title{Spatial-Temporal Alignment Network for Action Recognition}
\shorttitle{STAN}
\author[1]{Jinhui Ye}
\author*[1,2]{Junwei Liang}
\address[1]{AI Thrust, The Hong Kong University of Science and Technology (Guangzhou)}
\address[2]{Department of Computer Science and Engineering, The Hong Kong University of Science and Technology}
\fcssetup{
  received       = {month dd, yyyy},
  accepted       = {month dd, yyyy},
  corr-email     = {junweiliang@hkust-gz.edu.cn},
}

\newcommand{\fancyname}{\textit{STAN}}

\usepackage{cleveref}
\crefname{section}{\S}{\S\S}
\Crefname{section}{\S}{\S\S}

\newcommand{\eat}[1]{} % ignores argument

\begin{abstract}
This paper studies introducing viewpoint invariant feature representations in existing action recognition architecture. 
Despite significant progress in action recognition, efficiently handling geometric variations in large-scale datasets remains challenging. 
To tackle this problem, we propose a novel Spatial-Temporal Alignment Network (\fancyname), which explicitly learns geometric invariant representations for action recognition. 
Notably, the {\fancyname} model is light-weighted and generic,
which could be plugged into existing action recognition
models (e.g., MViTv2) with a low extra computational cost.
We test our {\fancyname} model on widely-used datasets like UCF101 and HMDB51. The experimental results show that the {\fancyname} model can consistently improve the state-of-the-art models in action recognition tasks in trained-from-scratch settings.
\end{abstract}
\fussy

\begin{document}

\section{Introduction}
\label{sec:intro}
% 0. define problem
% Human vision has the remarkable ability to recognize video actions efficiently, regardless of viewpoint variations.

Action recognition is a critical task in computer vision, aiming to identify specific actions in videos. These tasks are challenging due to the high variability in the appearance, motion, viewpoint, and scale of actions across different videos. 
% 1. conv filter approach
Convolutional neural networks (CNNs) \cite{cnn-lecun-eccv10,cnn-dtran-iccv15,cnn-Carreira-Zisserman-cvpr17,DBLP:journals/corr/SigurdssonDFG16,DBLP:conf/nips/FeichtenhoferPW16} 
have been leveraged to their full potential, employing spatial-temporal filters on GPUs/TPUs to recognize actions, which
outperforms traditional models including oriented filtering in space time (HOG3D) \cite{DBLP:conf/bmvc/KlaserMS08}, spatial-temporal interest points \cite{DBLP:conf/cvpr/LaptevMSR08}, motion history images \cite{DBLP:journals/tsmc/TianCLZ12}, and trajectories \cite{DBLP:conf/iccv/WangS13a}.
However, the art of action recognition is still far from satisfactory, compared with the success of 2D CNNs in image recognition \cite{he2016deep}.
Recently, vision transformers including ViT~\cite{dosovitskiy2020image} and MViT~\cite{fan2021multiscale}, which are grounded in the self-attention mechanism, have been introduced to address challenges in image and video recognition. These models have demonstrated noteworthy performance. Unlike CNNs that model pixels, transformers concentrate on visual tokens through attentions.
The inductive bias of translation invariance 
in CNNs makes it require less training data than 
attention-based transformers in general. 
Conversely, transformers present a unique strength in better leveraging large datasets, resulting in improved accuracy compared to CNNs.
Hence, transformers combined with low-level convolutional operations have been proposed \cite{fan2021multiscale,li2022mvitv2,liang2022multi} to further improve the 
efficiency and accuracy.

% 2.  key challenge: variations/viewpoint change
However, a key challenge of action recognition is to capture the variations across space and time.
%Since CNN assumes the filters share weights at different locations, 
But CNNs and transformers can not explicitly model the viewpoint changes and other variations. 
This is in contrast to human action recognition, which can efficiently recognize video actions, regardless of viewpoint variations.
% how to solve the challenge
%To overcome these limitations, the practical approach is to expand feature representations to provide a higher degree of freedom.  
To overcome these limitations, the practical approach is to design a model that could mimic human vision by explicitly countering viewpoint changes over time~\cite{liang2020spatial}.
%For example, the two-stream network \cite{twostream-Simonyan-nips14} proposes to integrate optical flow with RGB features. More recently, SlowFast
%\cite{feichtenhofer2019slowfast} combines both slow and fast pathways to learn different temporal information and obtain reasonable performance. However, such feature expanding approaches can quickly lead to
%cumbersome and high-dimensional feature maps. This not only increases computational costs but also overlooks the geometric interpretation of the subjects. 
% ours
% In this paper, 
% we show that our method can improve on both simpler backbone networks like ResNet3D and more complex ones like SlowFast, with minimal computation overhead.

% \begin{figure}[t!]
% \centering
% \includegraphics[width=0.47\textwidth]{figures/STAN_w_troi2.pdf}
% \caption{XXX XXX XXX XXX XXX XXX}
% \label{fig:title}
% \end{figure}

% 3. ours
To address the aforementioned challenges, we propose \fancyname, to tackle the viewpoint variation challenge in action recognition. Instead of stacking deeper CNN or transformer layers, {\fancyname} aims to learn geometric transformations and viewpoint invariant features explicitly and efficiently. 
The idea is motivated by \cite{kosiorek2019stacked}, which believes that  human vision relies on coordinate frames. 
However, the stacked capsule autoencoder in \cite{kosiorek2019stacked} is designed for 2D images, and too expensive for large-scale video action recognition.

\begin{figure*}[ht]
	\centering		\includegraphics[width=0.95\textwidth]{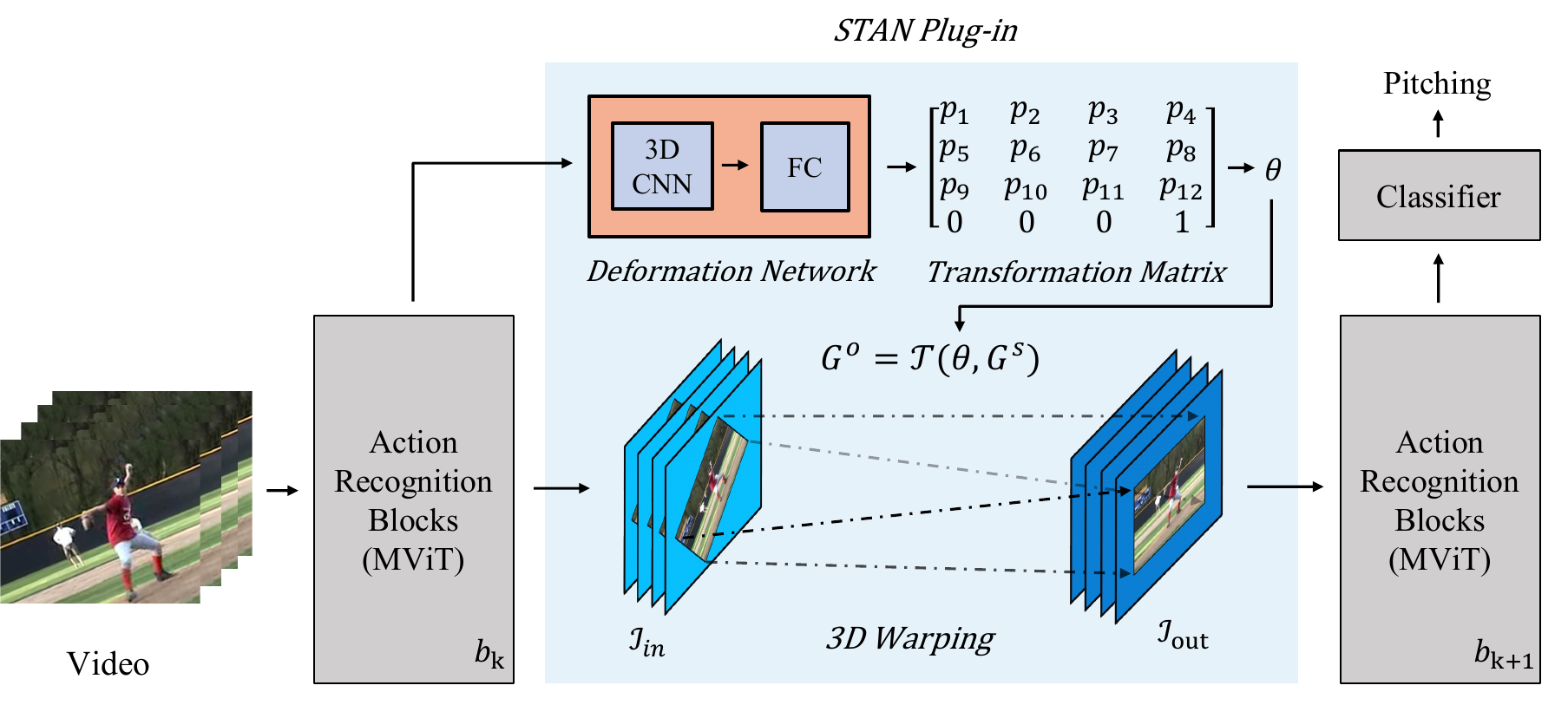} 
	\caption{The design of the Spatial Temporal Alignment Network ({\fancyname}). This figure illustrates the integration of {\fancyname} into an action recognition backbone network, such as MViTv2. Specifically, we highlight the two key components of the {\fancyname} plug-in: the Deformation Network and the Warping Module. The Deformation Network generates the geometric alignment parameter \( \theta \), while the Warping Module uses \( \theta \) to transform the input feature maps, thereby producing the final spatial-temporal alignment feature maps, which are then fed back into the action recognition network and contribute to better recognition performances. 
 % \jinhui{done: detail introduction of STAN framework}
 }
\label{fig:overview}
\end{figure*}

 As illustrated in Fig.\ref{fig:overview}, {\fancyname} generates a spatial-temporal transformation matrix $\theta \in \mathbf{R}^{4 \times 4}$ to warp the input feature map by taking the original spatial-temporal input tensor and extracting viewpoint invariant hints. This process produces an output tensor of the same shape as the original input tensor, but with features that are more robust to viewpoint variations. {\fancyname} is lightweight and generic, which allows it to be easily integrated into existing action recognition models like SlowFast~\cite{feichtenhofer2019slowfast}, and MViTv2~\cite{li2022mvitv2}, providing a more efficient and effective solution for action recognition.

% show improvement
% As discussed in Section~\ref{sec:arch},
%As shown in Table~\ref{tab:main_results}, 
As illustrated in Table~\ref{tab:main_results}, the integration of the {\fancyname} plugin into the MViTv2 structure significantly enhances model performance. Specifically, on the UCF101 dataset, we observe a relative increase of 3.84\% in accuracy, while the computational overhead, quantified in FLOPs, only experiences a slight rise of 1.57\% (from 64.39 G FLOPs to 65.40 G FLOPs). Similarly, on the HMDB51 dataset, the model's accuracy witnesses a substantial improvement of 6.4\%, with an almost negligible increment in computational cost. 
These results highlight the effectiveness of our {\fancyname} plugin in enhancing the performance of the MViTv2 architecture across different datasets.

% These enhancements come at a reasonable computational expense, underscoring the efficiency of our {\fancyname} plugin.

% 5. contribution and novelty
This paper makes the following significant contributions:

(1) We pioneer the exploration of explicit spatial-temporal alignment in 3D CNNs and transformer-based architectures for action recognition. This novel approach allows us to learn geometric invariant representations explicitly.

(2) We introduce the {\fancyname} model, a versatile and efficient solution that can be seamlessly integrated into existing action recognition architectures. The model is designed to add minimal computational overhead, making it a practical choice for enhancing existing model performance.

(3) We provide extensive experimental evidence across multiple datasets to demonstrate that the {\fancyname} model consistently improves upon the performance of existing state-of-the-art models in action recognition tasks. This robust performance underscores the effectiveness of our approach and its potential to contribute to the field of action recognition.

\section{The {\fancyname} Model}
\label{sec:method}

In this section, we describe our spatial-temporal alignment network for action recognition, which we call {\fancyname}.
% Motivated by the previous  works in image understanding \cite{DBLP:conf/nips/HuangMLL12,jaderberg2015spatial,lin2017inverse}, our work considers a generalized model in the video domain so that it can handle dynamic viewpoint changes in action recognition and detection tasks.
The key idea of {\fancyname} is to learn an explicit spatial-temporal transformation 
% spatial-temporal alignment 
for feature maps, accounting for viewpoint changes and actor movements within the videos. 
% A significant benefit of this approach is that it can achieve effects such as camera stabilization (Shown in Fig.\ref{fig:overview}), which is particularly crucial in Internet videos to counteract camera motions.

% The learned affine transformation (Eqn.~\ref{eqn:affine}) at
%  can achieve effects, including
% camera stabilization (Shown in Fig.~\ref{fig:overview}), which is important in Internet videos to counter camera motions.
Formally, given an input spatial-temporal feature map $\mathcal{I}_{in} \in \mathbf{R}^{C \times T \times H \times W}$ where $H$ stands for height, $W$ for width, $T$ for time and $C$ for channels, the {\fancyname} alignment function is defined as
\begin{align}
\label{eqn:stan}
    \mathcal{I}_{\text{out}} =
    \fancyname
    (\mathcal{I}_{\text{in}}) &= \mathcal{T}(\theta, \mathcal{I}_{\text{in}}) + \mathcal{I}_{\text{in}}, \\
    \text{where} \; \theta &= \mathcal{D}(\mathcal{I}_{\text{in}})
\end{align}

% \begin{align}
% \label{eqn:stan}
%     \text{STAN}(\mathcal{I}_{in}) = \mathcal{T}(\theta, \mathcal{I}_{in}) + \mathcal{I}_{in},
%     \text{where} \; \theta = \mathcal{D}(\mathcal{I}_{in})
% \end{align}

\eat{The output feature map $\mathcal{I}_{out}$ has the same number of channels as the input but could have different sizes in the temporal and spatial dimensions.
In this paper, we only explore the same-size transformation.
}

In this context, the output of {\fancyname} function, i.e., $\mathcal{I}_{out}$, is of the same dimension as $\mathcal{I}_{in}$. 
The function $ \mathcal{D}$ represents the 
deformation network~(\cref{sec:deformnet}), 
which computes the feature map alignment parameter $\theta \in \mathbf{R}^{4 \times 4}$ to warp the input feature map.
It can be in the form of a simple feed-forward network where the input is spatial-temporal features, and the output is alignment parameter $\theta$.

The function $ \mathcal{T}$ is defined as the 3D warping function~(\cref{sec:warp}), where input feature maps are warped based on the alignment parameter $\theta$.
In this paper, we add a residual connection in Eqn.~\ref{eqn:stan}  between the input feature maps and output feature maps for faster training and avoiding the boundary effect described in ~\cite{lin2017inverse}.

% The {\fancyname} layer can be added to different locations of the backbone to account for the alignment needs for different levels of feature maps.

% TODO detail construction
\subsection{Network Architecture}
\label{sec:Arch}
\noindent 
The {\fancyname} layer is designed as a plug-and-play module that can be added to different locations within the backbone network to account for the alignment needs of different levels of feature maps.
The overall case of {\fancyname}  is shown in Fig.~\ref{fig:overview}.
% In this paper, we use a 3D CNN as the backbone network to extract spatial-temporal feature maps from video frames. In addition,  {\fancyname}
% has the following key components:
{\fancyname} is designed to be integrated into existing action recognition backbone networks, such as MViTv2, which is currently the state-of-the-art transformer-based model for action recognition.
%renowned for extracting spatial-temporal feature maps from video frames and performing action recognition tasks. 

Within this architecture, {\fancyname} has two key components: The Deformation Network and the Warping Module. 
The Deformation Network produces the alignment parameter $\theta$ as defined in Eqn.~\ref{eqn:stan} to guide the deformation of spatial-temporal feature. 
The Warping Module transforms the input feature maps based on $\theta$ and outputs the final transformed spatial-temporal alignment  feature maps. Notably, this module acts as a sampler and does not contain any trainable parameters.
In the following, we will introduce the above modules in detail.

% Component one: Deformation Network
\subsection{Deformation Network}\label{sec:deformnet}
The deformation network $\mathcal{D}$ produces a transformation parameter, $\theta \in \mathbf{R}^{4 \times 4}$.
Our network is based on R(2+1)D~\cite{tran2018closer} although other options are possible, such as a simple feed-forward network, or a recurrent network and compositional function as proposed in ~\cite{lin2017inverse}.
Our deformation network generally consists of multiple CNN layers, followed by a fully-connected (FC) layer which outputs the transformation parameters. The dimension of the final ``FC'' layer depends on the type of parameterization we choose for the spatial-temporal alignment.

% All convolution layers are followed by a ReLU~\cite{hahnloser2000digital,glorot2011deep}. The dimension of the final ``FC'' layer depends on the type of parameterization we choose for the spatial-temporal alignment.
The design of {\fancyname} is flexible and can accommodate multiple types of transformation.
In this section, we introduce the affine transformation (Eqn.~\ref{eqn:affine}) as a method of parameterizing the deformation network, which has 12 degrees-of-freedom (DoF). The network output (\(\mathbf{P}_{\text{Aff}}=[p_1 \; ... \;p_{12}]^{T}\)) is of size 12, i.e., \(DoF = 12\), and \(\theta\) is constructed as 

\begin{align}\label{eqn:affine}
    \theta(\mathbf{P}_{\text{Aff}}) = 
    \begin{pmatrix}
    1 + p_1 & p_2 & p_3 & p_4\\
    p_5 & 1 + p_6 & p_7 & p_8\\
    p_9 & p_{10} & 1 + p_{11} & p_{12}\\
    0 & 0 & 0 & 1\\
\end{pmatrix}
\end{align}

% \junwei{ the p\_affine and p\_Aff should be consistent}
% \jinhui{done}
Other types of transformations, such as attention and homography transformations, are further discussed in Section~\ref{sec:ablation}. 
%The key intuition of the deformation on CNN feature maps is to compensate for the fact that CNNs are not rotation, scale, and shear transformation equivariant~\cite{kosiorek2019stacked}.

\subsection{Warping Module}
\label{sec:warp}
After computing the transformation matrix $\theta$ with the deformation network $\mathcal{D}$, we utilize a differentiable warping function $\mathcal{T}$ to transform the feature maps with better alignment of the content.
As shown in overview Fig.~\ref{fig:overview}.
the warping function is essentially a resampling of features from the input feature maps $\mathcal{I}_{in}$ to the output feature maps $\mathcal{I}_{out}$ at each corresponding pixel location.
Note that the feature maps could also be raw video frames.
Extending from the notation of 2D alignment~\cite{jaderberg2015spatial}, we define the output feature maps $\mathcal{I}_{out} \in \mathbf{R}^{C \times T \times H \times W}$ to lie on a spatial-temporal regular grid $G^o = \{G_i\}$, where each element of the grid $G_i = (t^o_i, x^o_i, y^o_i)$ corresponds to a vector of output features of size $\mathbf{R}^{C}$.  

Here we use affine transformation as an example, where the deformation matrix $\theta$ is parameterized by  $\mathbf{P}_{\text{Aff}}$ (Eqn.~\ref{eqn:affine}).
Given the coordinates mapping, 
%as the computed corresponding coordinates in the input feature maps might not be integers, 
we utilize the differentiable trilinear interpolation to sample input features from the eight closest points based on their distance to the computed point ($t^s_i, x^s_i, y^s_i$).
In this way, we iterate through every point in the regular grid, ($t^o_i \in [1, ..., T], x^o_i \in [1, ..., W], y^o_i \in [1, ..., H]$) and compute the output feature maps identically for each channel.

Hence, the pointwise sampling between the input and output feature maps is written as
\begin{align}
    \begin{bmatrix}
    t^o_i \\
    x^o_i \\
    y^o_i \\
    1 \\
    \end{bmatrix}
    =
    \begin{pmatrix}
    1 + p_1 & p_2 & p_3 & p_4\\
    p_5 & 1 + p_6 & p_7 & p_8\\
    p_9 & p_{10} & 1 + p_{11} & p_{12}\\
    0 & 0 & 0 & 1\\
    \end{pmatrix}
    \begin{bmatrix}
    t^s_i \\
    x^s_i \\
    y^s_i \\
    1 \\
    \end{bmatrix}
\end{align}

where ($t^o_i, x^o_i, y^o_i$) are the output feature map coordinates in the regular grid and ($t^s_i, x^s_i, y^s_i$) are the corresponding input feature map coordinates for feature sampling. 
%$p_i \in \theta$. % this is not correct

\subsection{Integration into Existing Backbone}\label{sec:arch}

% The integration of the {\fancyname} module into an existing network architecture is a delicate process that requires careful consideration. 

% to Resnet-3D, SlowFast
In Section~\ref{sec:Arch} we have defined the {\fancyname} layer, we next discuss how to effectively add it into existing action recognition backbone networks. In this paper, we focus on designing {\fancyname} for the RGB stream, leaving the extension to optical flow for future work.
Recent works~\cite{wang2018non,feichtenhofer2019slowfast,liang2022multi,pan2020actor,fan2021multiscale,li2022mvitv2} achieve single-stream state-of-the-art performance on action recognition with 3D CNNs and visual transformers. 
We, therefore, explore adding {\fancyname} layers to the widely used MViTv2~\cite{li2022mvitv2} network, which leverages the strengths of both 3D CNNs and transformers.

Intuitively, placing {\fancyname} layer at shallower layers allows earlier feature alignments that could potentially lead to better action recognition performance. 
However, shallow layers may not have enough abstraction in the feature maps for {\fancyname} to learn the right alignments.
Balancing this trade-off, we experiment with adding a {\fancyname} layer before different stages in the MViTv2 network.
This approach allows us to explore the impact of {\fancyname} at different levels of feature abstraction. The effectiveness of these insertion locations is further explored in the ablation experiments (\cref{sec:ablation}).

% \subsection{Detection Architecture}\label{sec:detect}

% \subsection{Training of {\fancyname}  in Architecture}\label{sec:detect}
% \junwei{this whole sub-section is argumentative and I suggest removing it}
% \jinhui{Removing Training of {\fancyname}  in Architecture, done}

% The {\fancyname} layer is designed as a plug-and-play module, and its application is not limited to a specific architecture. Still, it can be flexibly added wherever the alignment of spatial-temporal features is beneficial.
% The training process with {\fancyname} follows the standard procedures of the backbone network. The only difference is that during the forward pass, the {\fancyname} layer processes the feature maps, applying the learned spatial-temporal affine transformation $\theta$. The backward pass and parameter updates also include the parameters of the {\fancyname} layer, allowing it to continuously improve its alignment capability during training.

% In the context of action recognition, the final outputs of the MViTv2, after being processed by the {\fancyname} layer, are passed through global averaging and a fully-connected layer to obtain the action class probabilities. The integration of {\fancyname} ensures that the features used for recognition are not only spatially but also temporally aligned, leading to more accurate and robust action recognition results.

\section{Experiments}
\label{sec:exp}
To demonstrate the efficacy of our {\fancyname} module, 
%specifically, the viewpoint-invariant design that helps action models to generalize, 
we experiment on two popular action recognition datasets, UCF101~\cite{soomro2012ucf101} and HMDB51~\cite{kuehne2011hmdb}.
%In these experiments, we use the widely-used action recognition framework, i.e., MViTv2~\cite{li2022mvitv2}, as an example. 
We use the state-of-the-art MViTv2~\cite{li2022mvitv2} as our backbone model for all the experiments.
We aim to showcase that {\fancyname} can significantly enhance the performance of existing backbone networks in action recognition with only minimal computational overhead.

\subsection{Experimental Setup}
\label{sec:exp_rec}
%The action recognition task is defined to be a classification task given a trimmed video clip. To evaluate the generalization abilities of our proposed model, we consider two major action recognition datasets, UCF101 and HMDB51.

\noindent\textbf{Datasets.}
UCF101~\cite{soomro2012ucf101} is a widely used dataset for human action recognition. It consists of 13,320 videos from 101 action categories.
%, with each category containing 25 groups of videos. 
The videos are collected from YouTube and feature a variety of real-world challenges such as camera motion, diverse lighting conditions, and occlusions. 

HMDB51~\cite{kuehne2011hmdb} is another widely used dataset for action recognition. It comprises 6,766 video clips across 51 action categories, each with at least 101 clips. The videos, sourced from various internet platforms and digitized movies, are annotated with rich metadata, including occlusions, the number of people involved in the action, and camera movement. 
% This characteristic of HMDB51 makes it an excellent testing ground for the STAN module, particularly in its ability to handle viewpoint changes due to camera motion.

\noindent\textbf{Evaluation.} 
Our evaluation aims to assess the effectiveness and efficiency of the  {\fancyname} module in improving action recognition performance. To this end, we report action recognition accuracy (Acc), following the convention of previous works~\cite{carreira2017quo}. 
We run each experiment three times and show the averaged results. Additionally, we measure the computational cost by reporting 
% the inference time, 
the number of floating-point operations (FLOPs), and the number of parameters (Params). FLOPs are based on processing uniformly sampled $K=16$ clips from a video, each scaled to 224x224 from the original resolution.

\noindent\textbf{Baseline.} 
To validate the efficacy of our proposed {\fancyname} plug-and-play module, we employ the widely recognized MViTv2 architecture, a prevalent choice for action recognition tasks, as our baseline. Unless otherwise specified, we utilize the MViTv2 Small configuration as our primary model. This configuration comprises four stages, each characterized by a unique combination of channel width, the number of MViT blocks, and the number of heads in each block. Specifically, the channel widths are set as [96, 192, 384, 768], the number of MViT blocks are arranged as [1, 2, 11, 2], and the number of heads in each block is designated as [1, 2, 4, 8].lease refer to the original MViTv2~\cite{li2022mvitv2} paper.

\noindent\textbf{Implementation Details.}
In our experiments, we incorporate a single {\fancyname} layer into the MViTv2 architecture, as outlined in Section~\ref{sec:arch}. We adhere to the training recipe and inference strategies from MViTv2~\cite{li2022mvitv2} for Kinetics action classification.
%, but we switch the dataset to UCF101 and HMDB51. 
The strategy for handling the input clip and the spatial domain are identical to those in MViTv2.

Regarding the {\fancyname} module, we conduct experiments by inserting it after the second stage of the MViTv2 architecture, specifically between the third and fourth MViT blocks, echoing the block arrangement of [1, 2, 4, 8] mentioned earlier.
The default scaling of the {\fancyname} module is set to "Medium" in accordance with Table~\ref{tab:deformnet}, and the Transformation Mode is defined as an affine transformation with degrees-of-freedom (DoF). 
These choices are based on our understanding of {\fancyname} and the need to balance computational cost and performance. 
It's important to note that we did not perform an exhaustive search due to resource constraints.
More details can be found in the ablation study (\cref{sec:ablation}).

%Due to computational resource constraints, 
Our training is performed on 2 RTX 4090 GPUs with a batch size of 64. 
The training is conducted from scratch, without any pre-training.

% For the input clip and the spatial domain, we follow the same procedures as in MViTv2. We randomly sample a clip (T frames with a temporal stride of $\tau$) from the full-length video during training, and apply Inception-style cropping in the spatial domain.

% During inference, we uniformly sample $K=16$ clips from a video and scale the shorter spatial side to 256 pixels, taking a 224×224 center crop or 3 crops of 224×224 to cover the longer spatial axis. The final score is averaged over all predictions. 

\subsection{Main Results}
\label{sec:mainResults}
We evaluate the effectiveness of our proposed {\fancyname} module using the MViTv2 backbone on the UCF101 and HMDB51 datasets. The primary results are presented in Table~\ref{tab:main_results}.

% \begin{table}[]
% \centering
% \begin{tabular}{l|c|l|l}
%     \toprule

%         Models              & Acc             & FLOPs ($\Delta   \%$) & Params ($\Delta   \%$
% )       \\ \hline

%  3D-ConvNet~\cite{tran2015learning}      &    51.6       &  -    &    79   \\ 
% 3D-Fused~\cite{tran2015learning} & 69.5 & & 39 \\
%  Two-Stream$^*$~\cite{simonyan2014two} & 83.6  & - & 12 \\
%  Two-Stream I3D$^*$~\cite{tran2015learning} & 88.8 &- & 25 \\
% \hline
% MViTv2~\cite{li2022mvitv2}             &  84.97   &    64.39  & 34.31 \\
% % MViTv2 + {\fancyname}[0]             &  87.00   & \textbf{97.31} &    64.49 \\
%  ~~+ {\fancyname}[3] & \textbf{88.26} &  65.40 (1.57\%) & 34.56 (0.73 \%) \\ 

% \bottomrule
% \end{tabular}
% \caption{
% Experiment results on UCF101 dataset. We compare recent models with MViTv2 backbone as well as some classic methods. The metrics include accuracy (Acc), computation cost (FLOPs, in billions), and the number of parameters (Params, in millions). FLOPs is based on processing uniformly sampled $K=16$ clips from a video, each scaled to 224x224 from the original resolution. Models marked with $^*$ use optical flow information for action recognition.
% }

% \label{tab:ucf101}
% \end{table}

\noindent\textbf{UCF101.} As demonstrated in Table~\ref{tab:main_results}, our method significantly improves the performance of baseline. Specifically, when compared to the MViTv2 backbone, our method MViTv2 + {\fancyname}[3] improves the accuracy by 3.29, while the relative increase in computation cost is approximately 1.57\%.

\noindent\textbf{HMDB51.} 
As shown in Table~\ref{tab:main_results}, our proposed method also exhibits significant improvement over the baseline on the HMDB51 dataset. Specifically, the MViTv2 + {\fancyname}[3] model enhances the accuracy by 6.4\% compared to the MViTv2 backbone. The relative increase in computational cost is approximately 1.6\%, demonstrating the efficiency of our method.

% \begin{table}[]
% \centering
% \begin{tabular}{l|c|l|l}
%     \toprule

%         Models              & Acc             & FLOPs ($\Delta   \%$) & Params ($\Delta   \%$
% )       \\ \hline

%  3D-ConvNet~\cite{tran2015learning}      &    24.3       &  -    &    79   \\ 
% 3D-Fused~\cite{tran2015learning} & 37.7 & & 39 \\
%  Two-Stream$^*$~\cite{simonyan2014two} & 47.1  & - & 12 \\
%  Two-Stream I3D$^*$~\cite{tran2015learning} & 62.2 &- & 25 \\
% \hline
% MViTv2            &   68.97  &  64.39 & 34.27  \\
% % MViTv2 + {\fancyname}[0]             &  87.00   & \textbf{97.31} &    64.49 \\
% ~~+ {\fancyname}[3] & 73.40 & 65.41(1.6\%) & 34.55(0.82\%)\\ 

% \bottomrule
% \end{tabular}

% \caption{Experiment results on the HMDB51 dataset include Top-1 accuracy, computation cost (in billions), and the number of parameters (in millions) }
% \label{tab:hmdb51}
% \end{table}

\begin{table*}[]
\centering
\begin{tabular}{l c c|c c}
    \toprule
     \multirow{2}{*}{\textbf{Models}} & \multicolumn{2}{c|}{\textbf{Accuracy ($\Delta   \%$)}} & \multicolumn{2}{c}{\textbf{Computation ($\Delta   \%$)}} \\
    \cline{2-5}
    & UCF101  & HMDB51   &  Params (M) &  FLOPs (G)\\ \midrule
    3D-ConvNet~\cite{tran2015learning} & 51.6  & 24.3  & 79 & 108  \\ 
    3D-Fused$^*$~\cite{tran2015learning} & 69.5  & 37.7  & 39& - \\
    Two-Stream$^*$~\cite{simonyan2014two} & 83.6 & 47.1  & 12 & -\\
    Two-Stream I3D$^*$~\cite{tran2015learning} & 88.8  & 62.2  & 25 & 216 \\
    \hline
    MViTv2~\cite{li2022mvitv2} & 84.97  & 68.97 & 34.31 & 64.39   \\
    ~~+ {\fancyname}[3] &88.26(3.9\%)  & 73.40(6.4\%) &  34.56(0.73\%) & 65.40(1.57\%) \\ 
    \bottomrule
\end{tabular}
\caption{
Experiment results on UCF101 and HMDB51 datasets. We compare recent models with the MViTv2 backbone as well as some classic methods. The metrics used for comparison include accuracy, the number of parameters (Params, in millions), and computational cost (FLOPs, in billions).  Models marked with $^*$ use optical flow information for action recognition. The Params and FLOPs values of the classic methods are approximations based on ~\cite{tran2015learning} and ~\cite{feichtenhofer2019slowfast}. FLOPs and Params are reported referencing UCF101, and the values for the HMDB51 dataset are close similarities.}
% \jinhui{ merge FLOPs and Params and Delta for ACC}
% \jinhui{Done}
% \jinhui{FLOPs and Params of other methods need double checking}

\label{tab:main_results}
\end{table*}

\section{Analysis}

% XXX XXX XXX XXX XXX XXX XXX XXX
% XXX XXX XXX XXX XXX XXX XXX XXX
% XXX XXX XXX XXX XXX XXX XXX XXX
% XXX XXX XXX XXX XXX XXX XXX XXX
\subsection{Ablation Experiments}\label{sec:ablation}

\begin{table}[t!]
\centering
\begin{tabular}{l|l|l}
\toprule

& Acc($\Delta \%$) & FLOPs($\Delta \%$)  
               \\
\midrule
Basline MViTv2              & 84.97 &   64.39     \\ 
\midrule
% + {\fancyname} ([-1]) & - & 0.253 & 242.88 \\
\multicolumn{1}{c}{} & \multicolumn{2}{c}{ \em Insertion Position } \\
+ {\fancyname} ([0]) & 86.91(2.3\%) & 65.99(2.5\%) \\
+ {\fancyname} ([1]) & 87.00(2.4\%) & 65.49(1.7\%) \\
+ {\fancyname} ([3]) & \textbf{88.26(3.9\%)} & \textbf{65.40(1.6\%)} \\
+ {\fancyname} ([14])  & 87.96(3.5\%) & 65.25(1.3\%) \\ \midrule
 \multicolumn{1}{c}{} & \multicolumn{2}{c}{ \em Deformation Mode } \\
+ {\fancyname} (Att, 6)  & 87.24(2.7\%) & 65.40(1.6\%)  \\
+ {\fancyname} (Aff, 12) &\textbf{88.26(3.9\%)} & \textbf{65.40(1.6\%)}\\
% + {\fancyname} (Hom, 15)  & - & -  \\ 
\midrule
\multicolumn{1}{c}{} & \multicolumn{2}{c}{ \em Deformation Scale } \\
+ {\fancyname} (Small)  & 87.54(3.0\%) & 65.40(1.6\%) \\
+ {\fancyname} (Medium)  & \textbf{88.26(3.9\%)} & \textbf{65.40(1.6\%)} \\
+ {\fancyname} (Large) & 86.50(1.8\%) & 96.88(50.5\%) \\ \midrule
\multicolumn{1}{c}{} & \multicolumn{2}{c}{ \em Transfer Ability } \\
+ {\fancyname} (fixed $\mathcal{D}$)  & 85.98(1.2\%) & 65.40(1.6\%)  \\
% + {\fancyname} (fixed $W_{\theta}$)  & 83.86(-1.3\%) & 65.40(1.6\%)  \\
\bottomrule
\end{tabular}

\caption{Ablation experiment results (on UCF101). All ablation experiments are consistent with those in Section \ref{sec:exp_rec}, except for the factors being tested. ``+ STAN ([1])" represents inserting STAN after the $1st$ MViT block; ``+ STAN (Aff, 12)" represents using STAN with an affine transformation that has 12 degrees of freedom; ``STAN (Medium)" represents using STAN with a medium scale deformation network, as shown in Table~\ref{tab:deformnet}.    ``fixed $\mathcal{D}$" represents training STAN parameters on the HMDB51 dataset and using them directly on the UCF101. }
\label{tab:ablation}
\end{table}

In this section, we perform ablation studies on the UCF101 dataset with the MViTv2 architecture as the backbone network.
% We run each ablation
% experiments three times and show the averaged results. 
% The differences between the same run are within 0.1\%.
To understand how action models can benefit from {\fancyname}, we explore the following questions:

\subsubsection{{Where to insert {\fancyname} layer?}}
As outlined in Section~\ref{sec:exp_rec}, the baseline MViTv2 network is composed of four stages with 16 MViT blocks. Each stage of the architecture encodes different levels of visual features, with shallower layers capturing low-level features like edges and patterns, and deeper layers encoding more abstract information.
To determine the optimal placement of the {\fancyname} layer, we conduct a series of experiments, inserting the {\fancyname} layer before different stages within the MViT-S network.

The results, as shown in Table~\ref{tab:ablation}, indicate that the optimal placement of the {\fancyname} layer is after the second stage (the third MViT block), yielding a 3.9\% accuracy improvement with a minimal 1.6\% increase in computational cost. 
Interestingly, the placement of the {\fancyname} layer does not significantly impact the computational cost. This highlights the importance of strategic {\fancyname} layer placement over computational cost considerations.

\subsubsection{What is the best parameterization for the deformation network?}

In Section~\ref{sec:deformnet}, we introduced the affine transformation (Eqn.~\ref{eqn:affine}), denoted as `(Aff, 12) with 12 degrees-of-freedom (DoF). In this section, we expand our discussion to include another parameterization: attention transformation (Att, 6).

The attention transformation, as defined in Equation~\ref{eqn:att}, is more restrictive and allows for cropping, translation, and scaling. The network output \(\mathbf{P}_{\text{Att}}=[p_1,...p_6]^T\) is a vector of size 6, i.e., \(DoF = 6\), and \(\theta\) is constructed as follows:

\begin{align}\label{eqn:att}
    \theta(\mathbf{P}_{\text{Att}}) = 
\begin{pmatrix}
1 + p_1 & 0 & 0 & p_4\\
0 & 1 + p_2 & 0 & p_5\\
0 & 0 & 1 + p_3 & p_6\\
0 & 0 & 0 & 1\\
\end{pmatrix}
\end{align}

The results, as shown in Table~\ref{tab:ablation}, show that the affine transformation (Aff, 12) outperforms the attention transformation (Att, 6), suggesting that more complex transformations can enhance action recognition performance.

\noindent\subsubsection{What is the optimal scale of deformation for the {\fancyname} network?}

The scale of deformation in the {\fancyname} network is a crucial factor. To study the effect of the scale of deformation, we experiment with three different scales: small , medium, and large. The network architectures and their corresponding number of parameters for each scale are illustrated in Table~\ref{tab:deformnet}.

\begin{table}[ht]
\centering
\begin{tabular}{l|l|l|c}
\toprule
\centering \textbf{Scale} & \textbf{Convs } & \textbf{Channels} & \textbf{Params(M)}\\
\midrule
Small & [2] & [$C//4$] & {0.25} \\
\hline
Medium & [3, 2] & [\(C\), $C//4$] & {0.28} \\
\hline
Large & [4, 3, 2] & [\(4*C\), \(C\), $C//4$] & {8.22} \\
\bottomrule
\end{tabular}
\caption{Deformation network architectures for different scales. Each row represents a scale: small, medium, and large. The columns show the convolutional layers (the numbers represent the cubic filter size, e.g., [2] refers to a \(2\times2\times2\) filter), the number of channels at each layer, and the parameter quantity, respectively. $C$ is the channels of the original input features.
Each convolution layer is followed by a max pooling layer with the same filter size. All models end with a global average pooling and a fully-connected layer (FC) with DoF output dimensions for constructing the transformation matrix $\theta$.
%\jinhui{Samll and Medium are too close}
}
\label{tab:deformnet}
\end{table}

The results, as shown in Table~\ref{tab:ablation}, indicate that the optimal deformation scale for the {\fancyname} network is the medium scale. This scale yields the best performance improvement of 3.9\% 

The small scale, while having a lower computational cost, does not provide as much accuracy improvement. On the other hand, the large scale significantly increases the computational cost by 50.5\%, but only improves the accuracy by 1.8\%.

\noindent\subsubsection{Does {\fancyname} Possess Transferability?}

In this section, we explore the transferability of the {\fancyname} model across different datasets. We first train the {\fancyname} model on the HMDB51 dataset and subsequently fine-tune the layers following the {\fancyname} layer on the UCF101 dataset.
As demonstrated in Table~\ref{tab:ablation} (under the ``fixed $\mathcal{D}$" ), the {\fancyname} model exhibits a notable improvement on the UCF101 dataset. This suggests that the deformation learned from one dataset can indeed be effectively transferred to another, highlighting the potential of {\fancyname} for cross-dataset action recognition tasks.

\subsection{Qualitative Analysis.}

\begin{figure}[t!]
	\centering
		\includegraphics[width=0.47\textwidth]{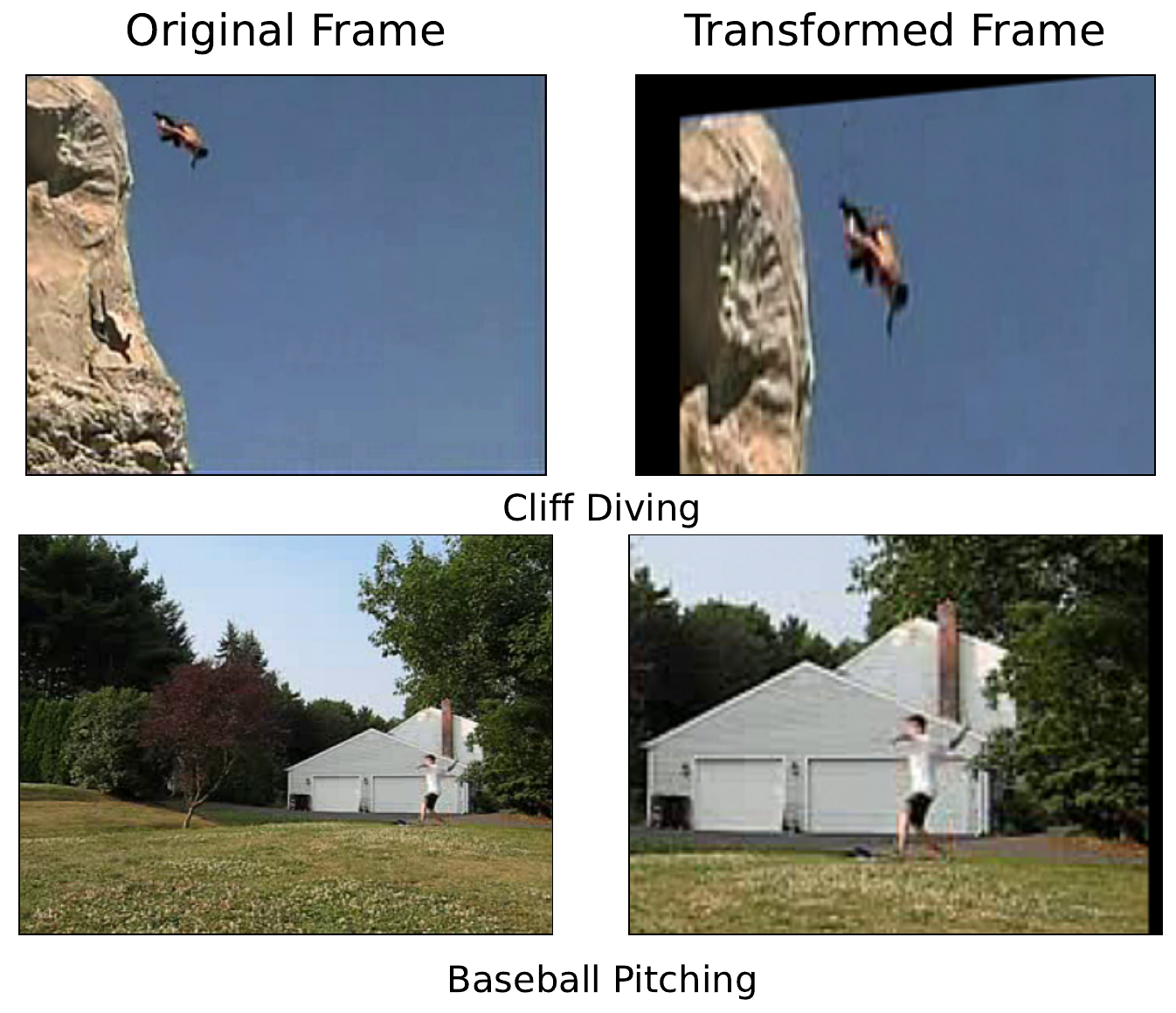} 
	\caption{Qualitative analysis. The left side column shows the original frames from the videos, and the right column shows the frames after the transformation by STAN. The top line corresponds to the action category 'Cliff Diving', where an actor is performing a cliff dive. The second row corresponds to 'Baseball Pitching', where an actor is pitching a baseball. STAN learns a transformation that centers the main actor. 
	}
	\label{fig:qual}
\end{figure}

In Section~\ref{sec:intro}, {\fancyname} can provide intuitive geometric interpretation of human actions. To illustrate such effects, we visualize example transformations of our model. 
We apply the output ($\theta$, Eqn.\ref{eqn:affine}) from the {\fancyname} layer to the original videos and illustrate the spatial transformation of the middle frames in the videos. As shown in Fig.\ref{fig:qual}, in the first scenario where the actor is cliff diving and is positioned at the edge of the video, {\fancyname} learns a transformation that centralizes the main actor. 
In case two, where an individual is pitching a baseball but the camera's viewpoint is quite distant, {\fancyname} learns a transformation that not only centralizes the actor but also magnifies the scene, thereby bringing the actor into a more prominent focus.

\section{Related Work}

The field of action recognition has seen significant advancements with the introduction of new datasets and models. The KTH dataset \cite{kth-icpr04}, one of the earliest benchmarks for action recognition, collected videos of individual actors performing six types of human actions (walking, jogging, running, boxing, hand waving, and clapping) against a clean background. However, due to the simplicity of these videos, the KTH dataset quickly became an easy benchmark as studies achieved near-perfect accuracy on it \cite{DBLP:conf/cvpr/CaoLH10,DBLP:conf/cvpr/WangKSL11}.

To address the limitations of the KTH dataset, the HMDB dataset \cite{hmdb51} was introduced in 2011, featuring 51 actions across 7000 video clips. The UCF101 dataset \cite{ucf101} further expanded this effort by collecting 101 action classes across 13000 clips. Both benchmarks were captured against more diverse backgrounds. Over the past decade, we have witnessed a steady improvement in accuracy on these two datasets through various methods, including feature fusion \cite{DBLP:journals/cviu/PengWWQ16}, two-stream networks \cite{twostream-Simonyan-nips14}, C3D \cite{cnn-dtran-iccv15}, I3D \cite{cnn-Carreira-Zisserman-cvpr17}, graph-based approaches \cite{wang2018videos,chen2019graph,qi2018learning,zhang2019structured}, and others \cite{xu2017r,chao2018rethinking,hou2017tube}. 
Recently, vision transformers such as ViT and MViT, founded on the self-attention mechanism, have emerged to tackle issues in image and video recognition, and they have showcased significant results. 
In contrast to CNNs that model pixels, transformers focus their attention on visual tokens. 
Many transformer-based methods have been proposed and proved to be effective in action recognition~\cite{li2022mvitv2,fan2021multiscale,liang2022multi}.

This paper focuses on action recognition using spatial-temporal alignment. In recent years, there has been a consistent effort to use alignment for image recognition \cite{DBLP:conf/nips/HuangMLL12,Xiong_2013_CVPR,jaderberg2015spatial,lin2017inverse,kosiorek2019stacked,dai2017deformable}. However, many previous studies have shown that alignment models are not as competitive as data-driven approaches like data augmentation or spatial pooling for image recognition. Some recent works have had to rely on very expensive models such as recurrent networks \cite{lin2017inverse} or stacked capsules \cite{kosiorek2019stacked}. As a result, many alignment-based recognition methods are limited to MNIST \cite{kosiorek2019stacked} and face recognition \cite{Xiong_2013_CVPR}. 
Some follow-up works on capsule networks~\cite{duarte2018videocapsulenet} and 2D spatial alignment networks~\cite{huang2019part,yang2019step} have been proposed.
Liang et. al.~\cite{liang2020spatial} have proposed to add an alignment network on 3D convolutional networks.
Deformable convolution~\cite{dai2017deformable}, a popular related work, computes spatial offsets to deform traditional convolution operations based on the visual content in images. 
Our method differs in that: (1) we aim for spatial-temporal alignment of features; (2) we explicitly compute geometric transformations and apply them on recent state-of-the-art transformer backbones. This paper shows that it is possible to build an efficient spatial-temporal alignment for action recognition, and improve recent networks with very few extra parameters.

\section{Conclusion}
\label{sec:concl}

This paper introduced the Spatial Temporal Alignment Network ({\fancyname}), a pioneering approach to action recognition. {\fancyname} is the first to explicitly incorporate spatial-temporal alignment in vision transformers for action recognition, leading to substantial performance enhancements.

The lightweight and adaptable design of {\fancyname} enables its seamless integration into existing models. We demonstrated its effectiveness using the state-of-the-art MViTv2 framework, underscoring the feasibility of an efficient spatial-temporal alignment network for action recognition.
Despite its simplicity, {\fancyname} significantly boosts the performance of these models, achieving remarkable accuracy improvements on the UCF101 and HMDB51 datasets with a minimal increase in computational cost.
% Our work paves the way for future research in viewpoint-invariant feature representation for actions. We anticipate that {\fancyname} will inspire further advancements in this field.

% \begin{acknowledgement}
% The ``Acknowledgements'' section is the general term for the list of contributions, credits, and other information included at the end of the text of a manuscript but before the references. Conflicts of interest and financial disclosures must be listed in this section. Authors should obtain written permission to include the names of individuals in the ``Acknowledgements'' section.
% \end{acknowledgement}

% \section*{Appendixes~(if needed)}

% \subsection*{Appendix A}

% \subsection*{Appendix B}

% \raggedbottom
\bibliographystyle{fcs}
\bibliography{ref}
% \vspace{48mm} % 将引用部分向下移动，以抵消前面的 \vspace{-48mm}
% \newpage 
% \raggedbottom
% \vfill
\vspace{-48mm} % 但是它同时破坏了\bibliography{ref} 的布局

\begin{biography}{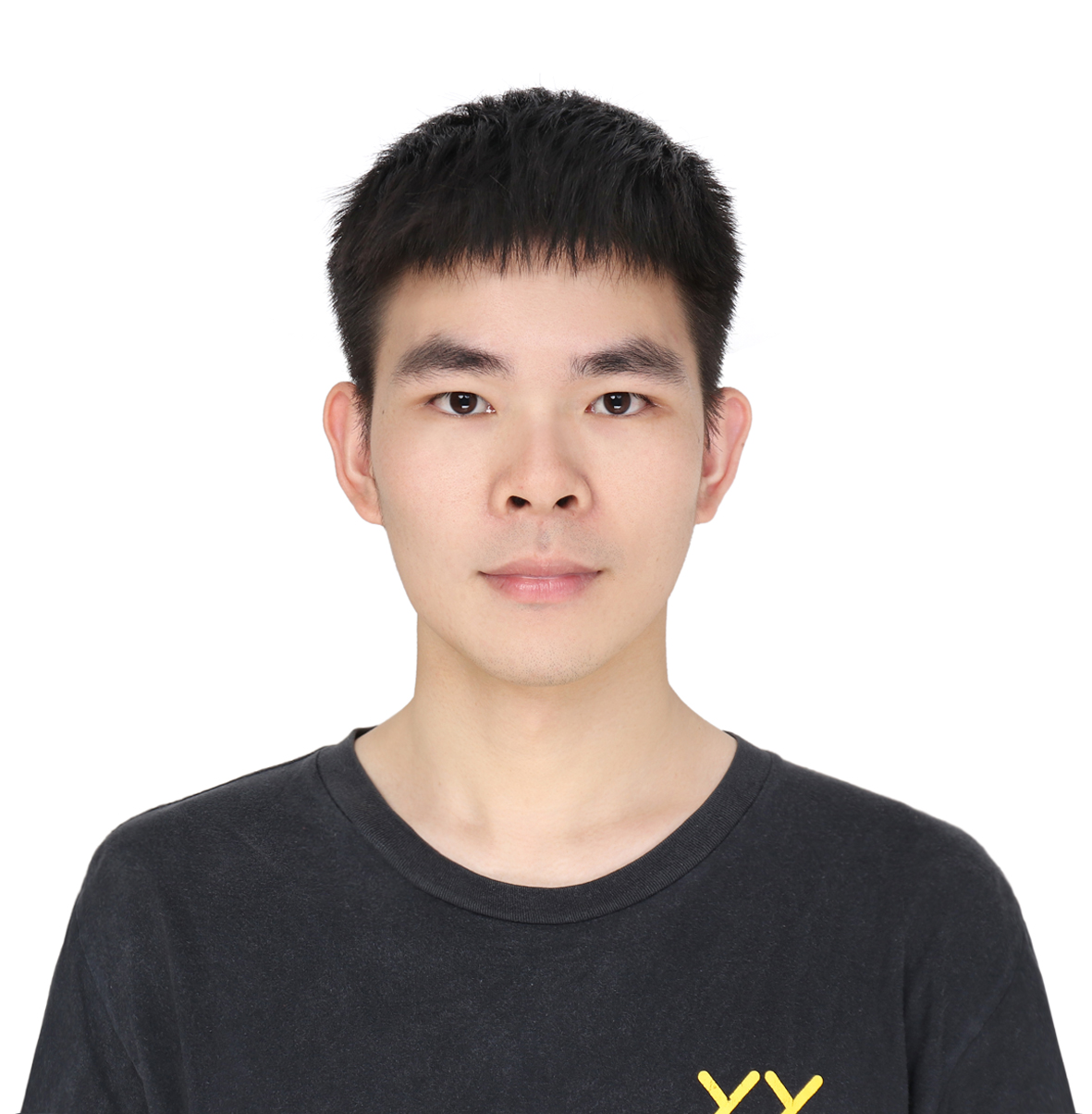}
Jinhui Ye is a Master of Philosophy candidate in Artificial Intelligence at The Hong Kong University of Science and Technology (Guangzhou). He is a dedicated researcher and AI enthusiast. He received his degree in Software Engineering from the South China University of Technology in 2022. 
His research interests lie in the intersection of gesture-based understanding, perception, and human-machine collaboration based on language. He previously interned at Tencent AI Lab, working on sign language translation. He is part of the Multimodal Human-Computer Interaction Group at HKUST-GZ, focusing on multimodal human-computer interaction and collaboration.
\end{biography}

\begin{biography}{figures/junwei_pic}
Dr. Junwei Liang is a tenure-track Assistant Professor (TTAP) at The Hong Kong University of Science and Technology (Guangzhou). He is also affiliated with HKUST CSE. He was a senior researcher at Tencent Youtu Lab. Before that, he received his Ph.D. from Carnegie Mellon University. He received the Baidu Scholarship, Yahoo Fellowship and ICCV Doctoral Consortium Award. He received the Rising Star Award at the World AI Conference in 2020. He is the winner of several public safety video analysis competitions, including NIST ASAPS and TRECVID ActEV. His work has helped and been reported by major news agencies like the Washington Post and New York Times. His research interests include human trajectory forecasting, action recognition, and large-scale computer vision.

\end{biography}

\end{document}